\documentclass[runningheads]{llncs}
\usepackage{graphicx}

\usepackage{multirow}
\usepackage{multicol}
\usepackage{lineno}
\usepackage{amsmath}
\usepackage{algorithmic}
\usepackage{booktabs}
\usepackage{caption}
\usepackage[inline]{enumitem}
\usepackage{algorithm}
\usepackage[algo2e]{algorithm2e} 

\usepackage{times}
\usepackage{epsfig}
\usepackage{amssymb}
\usepackage{subcaption}
\usepackage{amsfonts}
\usepackage{array}
\usepackage{textcomp}
\usepackage{makecell}
\usepackage{diagbox}
\usepackage{threeparttable}
\usepackage{mathtools}
\usepackage{eqparbox}
\usepackage{bm}
\usepackage{bbding}
\usepackage{comment}
\usepackage{color}
\usepackage{bbding}
\usepackage{pifont}
\usepackage{wasysym}
\usepackage{cite}
\usepackage{stfloats}
\usepackage{float}

\DeclareMathOperator{\x}{\mathbf{x}}

\begin{document}
\title{KV Inversion: KV Embeddings Learning for Text-Conditioned Real Image Action Editing\thanks{J. Huang and Y. Liu—Contributed
equally to this work. \\This work is supported by Key-Area Research and Development Program of Guangdong Province (2019B010155003), the Joint Lab of CAS-HK, and Shenzhen Science and Technology Innovation Commission (JCYJ20200109114835623, JSGG20220831105002004).}}
\titlerunning{KV Inversion: KV Embeddings Learning for Real Image Editing}
%
%
\author{Jiancheng Huang\inst{1,2} \and
Yifan Liu\inst{1} \and Jin Qin\inst{1,2} \and
Shifeng Chen\inst{1,2}\textsuperscript{(\Envelope)}}
\authorrunning{J. Huang et al.}
%
\institute{ShenZhen Key Lab of Computer Vision and Pattern Recognition, Shenzhen Institute of Advanced Technology, Chinese Academy of Sciences, Shenzhen, China \\\email{\{jc.huang,yf.liu2,shifeng.chen\}@siat.ac.cn} \and
University of Chinese Academy of Sciences, Beijing, China
}
\maketitle              
\begin{abstract}
   Text-conditioned image editing is a recently emerged and highly practical task, and its potential is immeasurable. However, most of the concurrent methods are unable to perform action editing, i.e. they can not produce results that conform to the action semantics of the editing prompt and preserve the content of the original image. To solve the problem of action editing, we propose KV Inversion, a method that can achieve satisfactory reconstruction performance and action editing, which can solve two major problems: 1) the edited result can match the corresponding action, and 2) the edited object can retain the texture and identity of the original real image. In addition, our method does not require training the Stable Diffusion model itself, nor does it require scanning a large-scale dataset to perform time-consuming training.

\end{abstract}
\section{Introduction}
Along with the widespread adoption of diffusion model and muti-modal generative model, recent months have witnessed remarkable progress in text-to-image generation, which is mainly used for generation in many real-world scenarios, such as AI painting, commercial design, film making, etc.~\cite{ramesh2021zero,nichol2021glide,yu2022parti,ramesh2022hierarchical,rombach2022high}. For example, Stable Diffusion~\cite{rombach2022high} is able to generate diverse and high quality images according to user-provided text prompts. However, it is not enough just to generate a completely new image, and it will be more exciting to generate the desired image given a editing prompt, which we call text-conditioned image editing. There are many methods that use a pre-trained large-scale text-to-image model to implement text-conditioned image editing~\cite{nichol2021glide,hertz2022prompt,tumanyan2022plug,parmar2023zero}.
Text-conditioned image editing is a recently emerged and highly practical task, and its potential for successful application in comic book production, video editing, advertising material production is immeasurable. 

Note that \textbf{synthetic image editing} and \textbf{real image editing} are two very different tasks. Synthetic image editing means that when an image is synthesised using a source prompt, an editing prompt can be used to create another image that is consistent with it, and the user cannot provide a real image in the process. Real image editing, on the other hand, means that the user can provide a real image and then use the edit prompt to edit that real image. We can divide this emerging field into 5 categories according to the content to be edited:
1) Object replacement, where the object in the original diagram is replaced with another object.
2) New objects creation, where a new object is added to the editing image and the rest remains unchanged.
3) Action editing, the object retains its original texture and identity, but the action becomes the new action given by the edit text.
4) Scene editing, where the object remains unchanged and the scene is turned into a given scene.
5) Style editing, the art style or colour style is changed.

\begin{figure*}[t]
    \centering
    \includegraphics[width=\linewidth]{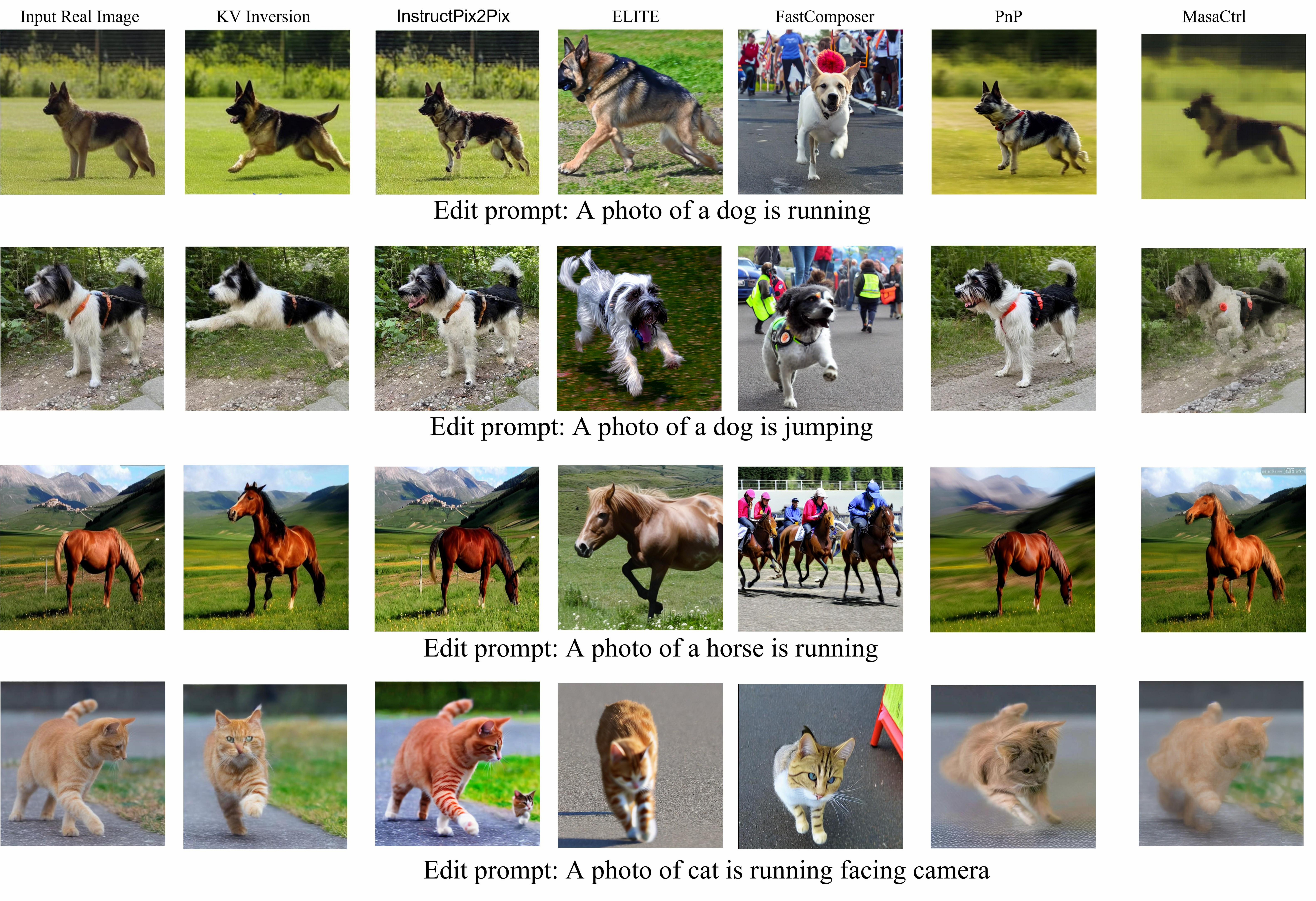}
    \vspace{-22pt}
    \caption{Comparing with different concurrent image editing methods on the real natural world images, it is obvious that the object in the editing result of our KV Inversion can meet the edit prompt corresponding to the action editing, while retaining the original real image object.}
    \label{fig:results_real}
    \vspace{-23pt}
\end{figure*}

 A number of existing text-conditioned image editing methods~\cite{hertz2022prompt,tumanyan2022plug,parmar2023zero} enable the above types of 1), 2), 4) and 5) to be edited. Most of them are able to perform tasks such as object replacement, new object creation and style transfer while keeping the overall structure and layout unchanged, and achieve satisfactory results. However, these methods are unable to perform type 3) edits as shown in Fig.~\ref{fig:results_real}. Some works aim to address this problem, such as Textual Inversion~\cite{gal2022image}, Imagic~\cite{kawar2022imagic}, DreamBooth~\cite{ruiz2022dreambooth}, Custorm Diffusion~\cite{kumari2022customdiffusion}, ELITE~\cite{wei2023elite}, FastComposer~\cite{xiao2023fastcomposer} and MasaCtrl~\cite{cao2023masactrl} are proposed to achieve the type 3) action editing. They can accomplish action editing while retaining the basic properties of the original image. However, Textual Inversion~\cite{gal2022image}, Imagic~\cite{kawar2022imagic}, DreamBooth~\cite{ruiz2022dreambooth} and Custorm Diffusion~\cite{kumari2022customdiffusion} require finetune of the diffusion model itself as well as the text embedding on 5 images of the editing object, so the GPU memory and time cost required is too high. ELITE~\cite{wei2023elite} requires training on a large dataset, which is even more expensive on time and GPU. MasaCtrl does not require training and finetuning, but its reconstruction performance on real images is unsatisfactory, making it difficult to editing.

 In our method, we aim to solve the problem of action editing without using multiple images of the same object for finetuning (generally known as \textbf{Tuning-free} in this field), without training the diffusion model itself, and without training on a large dataset for a long time (generally known as \textbf{Training-free} in this field), and to propose a solution to the mentioned challenges in the above setting. The core problem is how to preserve the content of the original object when the action changes. Our KV Inversion is divided into 3 stages. The fundamental difference with the concurrnet works~\cite{hertz2022prompt,parmar2023zero,chefer2023attend} is that KV Inversion directly learns Key and Value at the self-attention layer to preserve the content of the source image by these learnable parameters (KV embeddings), which we call upgrading the original self-attention to Content Preserving self-attention (CP-attn).
Then, in the editing stage, we use the edit text prompt to introduce the action information and the learned KV embeddings to preserve the texture and identity of the object. We further control the timesteps of upgrading the CP-attn and the segmentation mask obtained by Segment Anything (SAM)~\cite{kirillov2023segment}, thus achieving a faster and more controllable editing.

Our main contributions are summarized as follows. 1) We propose a training-free text-conditoned image action editing method KV Inversion to solve the action editing problem. 2) We design an upgrade version of self-attention named content preserving self-attention, which can preserve the texture and identity of the original object and then be used to fill the editing image. 3) Comprehensive Experiments show that KV Inversion can achieve satisfactory performance in real image editing. 

\begin{figure*}[t]
    \centering
    \includegraphics[width=0.99\linewidth]{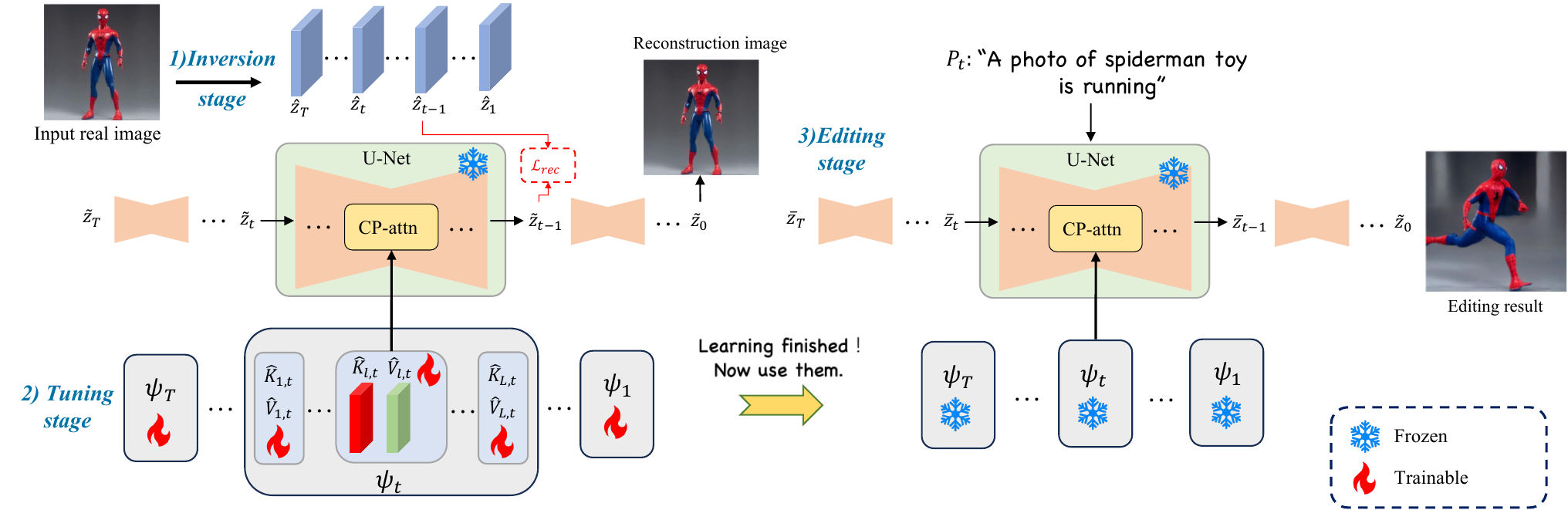}
    \vspace{-6pt}
    \caption{Pipeline of the proposed KV Inversion, which can be divided into 3 stages, inversion stage, tuning stage and editing stage. Inversion stage is for getting supervision. Tunning stage is for learning KV embedding for content preserving and editing stage is using the learned content for consistency editing results.} 
    \label{fig:main_arch}
    \vspace{-22pt}
\end{figure*}

\vspace{-9pt}
\section{Background}
\vspace{-5pt}
\subsection{Text-to-Image Generation and Editing}
 
Text-to-image generation models~\cite{song2019generative,sahariaphotorealistic,ramesh2022hierarchical,rombach2022high,yu2022scaling} have experienced an unprecedented surge in popularity. Initially, most early image generation methods relied on GANs~\cite{reed2016generative,zhang2018stackgan++,xu2018attngan,li2019controllable,zhu2019dm,zhang2021cross,tao2022df} and conditioned on text descriptions. These models align text descriptions with synthesized image contents using multi-modal vision-language learning.   However, more recently, diffusion models~\cite{song2019generative,ho2020denoising,nichol2021improved,dhariwal2021diffusion,patashnik2021styleclip,gal2022image,huang2023wavedm} have emerged as a dominant force, delivering exceptional generative power and achieving cutting-edge results in image quality and diversity. Notably, several text-to-image diffusion models, including DALL·E 2~\cite{ramesh2022hierarchical}, LDM~\cite{rombach2022high}, VQ-Diffusion~\cite{gu2022vector}, InstructPix2Pix~\cite{brooks2022instructpix2pix} and GLIDE~\cite{nichol2021glide}, have been developed to further enhance the synthesis process. The rise of diffusion-based models~\cite{patashnik2021styleclip,gal2022image} has showcased the potential for manipulating given images without human intervention.

Editing images based on text descriptions is a complex task.
Recently, there has been extensive research on text-conditioned editing using GANs~\cite{nam2018text,li2020manigan,xia2021tedigan,patashnik2021styleclip,RN20,karras2021alias,RN22,abdal2020image2stylegan++,RN32}. The introduction of CLIP~\cite{RN17} has revolutionized image editing by allowing people to guide the editing process with text inputs. Diffusion models \cite{ramesh2022hierarchical,saharia2022photorealistic} exhibit inherent capabilities for image editing due to their strong ability to extract text features using CLIP. An innovative approach called VQGAN-CLIP~\cite{crowson2022vqgan} combines VQGAN~\cite{esser2021taming} and CLIP~\cite{radford2021learning} in an auto-regressive model, enabling the production of high-quality images and precise edits with diverse and controllable outcomes. 

Through training, 
the diffusion model generates subject-specific images aligned with textual descriptions, opening new possibilities for precise image generation.

\vspace{-10pt}
\subsection{Stable Diffusion Model}
\vspace{-3pt}
Diffusion models~\cite{ho2020denoising,song2020denoising,nichol2021improved} utilize a forward process that gradually adds Gaussian noise to training data, and then recover the data distribution by a reverse process. The forward process follows the Markov chain that transforms a data sample $\x_0\sim q(\x_0)$ into a sequence of noisy samples $\x_{1:T}=\x_1, \x_2, \cdots, \x_T$ in $T$ steps. The above process can be reversed by $p_{\theta}(\x_{0:T})$ with learnable parameters $\theta$, where $p_{\theta}(\x_{0:T})$ is implemented by a neural network. The optimization of it can be converted to training a network $\bm{\epsilon}_{\theta}(\x_t,t)$ to predict the Gaussian noise vector added to $\x_t$~\cite{ho2020denoising}.

KV Inversion is designed for Stable Diffusion (SD)~\cite{rombach2022high}, whose contribution is learning the distribution of latent space data rather than image space. A autoencoder network is adopted in SD for the encoding and decoding and the U-Net~\cite{ronneberger2015unet} is as the noise predict network $\epsilon_\theta$ for latent noises $\{\mathbf{\hat{z}}_t\}$. The U-Net $\epsilon_\theta$ consists of many CNN block, self-attention layer and cross-attention layer, where the self-attention and cross-attention layers are responsible for the flow of information from the text prompt and the features themselves, respectively. Query features $Q$ is derived from the image features, while Key and Value features $K, V$ are derived from the text embedding (cross-attention) or the image itself (self-attention).  There are several phenomena: 1) The attention map of cross-attention determines the structure of the image (the position of the appearing objects) ~\cite{hertz2022prompt,tumanyan2022plug,chefer2023attend}. 2) $K$ and $V$ of the cross-attention layer change the texture and detail of generated image. 3) $K$, $V$ and the attention map in self-attention have a dramatic effect on the generated result in terms of content.
\vspace{-7pt}
\section{KV Inversion: Training-Free KV Embeddings Learning}
\vspace{-4pt}
\subsection{Task Setting and Reason of Existing Problem}
Given a real rather than a synthetic source image $I_s$ and a corresponding text prompt $P_s$ (which can be semantic text or empty), the goal of our task is to synthesize a new ideal image $I_t$ with a pretrain stable diffusion model that matches the target editing text prompt $P_t$. This editing result image $I_t$ should meet the following requirements: 1) $I_t$ semantically matches the text prompt of $P_t$, e.g., it can satisfy that the corresponding object is performing the corresponding action. 2) The object inside $I_t$ should be consistent with $I_s$ in terms of content. For example, given a real image (corresponding to $I_s$) with a cat standing still, we edit the text prompt $P_t$ to "a running cat", and then generate a new image with the cat running, while the other contents of the image remain basically unchanged.

This task is very difficult, especially when used on real images, and most of the current stble diffusion based image editing methods do not allow for action editing on real images while maintaining good reconstruction performance~\cite{hertz2022prompt,tumanyan2022plug}. A naive baseline method for action editing is to directly use $P_t$ to synthesize a new image $\bar{I}_t$ that matches the semantics of the actions and objects in $P_t$. However, even using $\mathbf{\hat{z}}_T$ as the starting point of the reverse process obtained from DDIM inversion of the original real image $I_s$, the scene and object generated in $\bar{I}_t$ are often very different from the original real image $I_s$~\cite{hertz2022prompt}.

The core problem mentioned above is that the scene and object generated in $\bar{I}_t$ are often very different from the original real image $I_s$. The primary cause of this phenomenon is that during editing, and the Key $K$ and Value $V$ of cross-attention in the U-Net are derived from the target editing text prompt $P_t$, which brings new content features different from those of the original real image $I_s$, so that the editing result has changed dramatically in content. Therefore, our core idea is to first implement a better inversion method to preserve the contents in the source image $I_s$ in the feature space. Then, we utilize these preserved content and combine it with the action layout generated by the target prompt $P_t$, and finally synthesize the desired editing image $I_t$.
\vspace{-8pt}

\subsection{KV Inversion Overview}
\vspace{-4pt}
To realise the above core idea, we propose KV Inversion, which can be divided into 3 stages, inversion stage, tuning stage and editing stage. 

\textbf{1) Inversion Stage: Getting $\mathbf{z}_t$ for Supervision.}
Specifically, as illustrated in Fig.~\ref{fig:main_arch}, for obtaining the trace of $\mathbf{\hat{z}}_t$ in the reserve process, we first perform DDIM inversion~\cite{dhariwal2021diffusion,song2020denoising} to synthesize a series of latent noises $\{\mathbf{\hat{z}}_t\}$ including $\mathbf{\hat{z}}_T$. However, directly using $\mathbf{\hat{z}}_T$ as the starting point of the reverse process can't reconstruct the original real image $I_s$ well since the U-Net can't predict the noise accurately enough which resulting in the next stage for better content preserving.
\vspace{-1pt}

\textbf{2) Tuning Stage: Learning KV Embeddings for Better Content Preserving.}
For better reconstruct $I_s$, we need more parameters to learn the content of $I_s$. Since we find that the contents (texture and identity) are mainly controled in the self-attention layer, we choose to learn the $K$ and $V$ embeddings in the self-attention layer. Here we use $\hat{K}_{l,t}$ and $\hat{V}_{l,t}$ to denote the learnable Key and Value embeddings in the $l$-th self-attention layer during step $t$ of reverse process. Then we define our proposed Content Preserving self-attention (CP-attn) as follows. For instance, given the $l$-th self-attention layer during step $t$ of reverse process, we have:
\begin{equation}
    \bar{K}_{l,t} = \lambda_{l,t}^k K_{l,t} + \gamma_{l,t}^k\hat{K}_{l,t},\quad
    \bar{V}_{l,t} = \lambda_{l,t}^v V_{l,t} + \gamma_{l,t}^v\hat{V}_{l,t},\\
\end{equation}
\begin{equation}
    \label{eq:attention}
    \text{Attention}(Q_{l,t}, \bar{K}_{l,t}, \bar{V}_{l,t}) = \text{Softmax}(\frac{Q_{l,t}{\bar{K}_{l,t}}^T}{\sqrt{d}})\bar{V}_{l,t},
\end{equation}
where $(K_{l,t},V_{l,t})$ and $(\bar{K}_{l,t},\bar{V}_{l,t})$ denote the original and the new (Key, Value), respectively. Besides,  $(\lambda_{l,t}^k,\lambda_{l,t}^v)$ and $(\gamma_{l,t}^k,\gamma_{l,t}^v)$ denote the learnable weights for the original and the learnable (Key, Value) embeddings.

\begin{algorithm}[t]
\SetAlgoLined
\textbf{Require:} the latent of the original real image $\mathbf{z}_0$ and the source prompt embeddings $\mathbf{c}_s$. The initialization learnable parameters $\psi=\{\psi_{t},\ t=1,...,T\}$.
 \vspace{1mm} \hrule \vspace{1mm}
\textbf{1) Inversion stage:} Set guidance scale $w=7.5$ of stable diffusion model, in this stage we utilize DDIM inversion to obtain the trace of noises $\{\mathbf{\hat{z}}_t, t=1,...,T \}$.
 \vspace{1mm} \hrule \vspace{1mm}
 \textbf{2) Tuning stage:} 
 Set guidance scale $w=7.5$ and the begining of reverse process $ \mathbf{\widetilde{z}}_T = \mathbf{\hat{z}}_T$; \\
 \For{$t=T,T-1,\ldots,1$}{
 \While{\textbf{not converge}}{
        obtain $\mathbf{\widetilde{z}}_{t-1}$ by $\epsilon_\theta(\mathbf{\widetilde{z}}_t,t,\mathbf{c}_s, \psi_t)$;\\
        $\psi_t  \leftarrow  \psi_t  - \eta \nabla_{\psi_t }\mathcal{L}_{rec}$;
 }

 obtain final $\mathbf{\widetilde{z}}_{t-1}$ by learned $\psi_t$ as Eq.~\ref{eq:pred_next};
 
 }
 \textbf{Return} learned parameters $\psi$
 \vspace{1mm} \hrule \vspace{1mm}
 \textbf{3) Editing stage:} 
 Set guidance scale $w=7.5$ and the begining of reverse process $ \mathbf{\bar{z}}_T = \mathbf{\hat{z}}_T$. Given a editing prompt $P_t$, then get its embedding $\mathbf{c}_t$ and an uncondition embedding $\mathbf{c}_u$;\\
 \For{$t=T,T-1,\ldots,1$}{
        $\epsilon_c=\epsilon_\theta(\mathbf{\bar{z}}_t,t,\mathbf{c}_t, \psi_t)$;\\
        $\epsilon_u=\epsilon_\theta(\mathbf{\bar{z}}_t,t,\mathbf{c}_u, \psi_t)$;\\
        $\epsilon_t=\epsilon_u + w(\epsilon_c - \epsilon_u)$;\\
        $\mathbf{\bar{z}}_{t-1} = \mathrm{Sample}(\mathbf{\bar{z}}_t, \epsilon)$;
 }
 $I_t = \mathrm{Decode}(\mathbf{\bar{z}}_0)$;\\
 \textbf{Return} Editing result $I_t$
\caption{The 3 Stages of KV Inversion}\label{alg:alg_twostage}
\end{algorithm}

Fig.~\ref{fig:main_arch} shows that our framework consists of the above learnable parameters ($\hat{K}_{l,t}$, $\hat{V}_{l,t}$, $\lambda_{l,t}^k$, $\lambda_{l,t}^v$, $\gamma_{l,t}^k$, $\gamma_{l,t}^v$) and the original diffusion networks, and we denote all the above learnable parameters with $\psi$ and the learnable parameters for step $t$ with $\psi_t$. 

In this tuning stage, we first use $\mathbf{\widetilde{z}}_{T}=\mathbf{\hat{z}}_{T}$ as the begining of reverse process ($T \rightarrow 1$). For each specific timestep $t$, we train the corresponding parameters $\psi_t$. We get the output of U-Net $\epsilon_\theta(\mathbf{\widetilde{z}}_t,t,\mathbf{c}_s, \psi_t)$ and use DDIM sampling to produce the next noise sample $\mathbf{\widetilde{z}}_{t-1}$, where $\mathbf{c}_s$ is the text embedding of source prompt $P_s$. Since the latent noise $\mathbf{\hat{z}}_{t-1}$ represents a reasonable trace whose corresponding result $\mathbf{\hat{z}}_0$ is close to the real image $I_s$, we use $\mathbf{\hat{z}}_{t-1}$ as supervision to train our learnable parameters $\psi_t$ to output a more precise noise $\mathbf{\widetilde{z}}_{t-1}$, which is closer to the noise representation $\mathbf{\hat{z}}_{t-1}$ than baseline DDIM inversion and other inversion methods~\cite{mokady2022null}. The loss function is 
\begin{equation}\label{eq:recon}
 \mathcal{L}_{rec} = \min_{\psi_t}  \left \|\mathbf{\hat{z}}_{t-1}-\mathbf{\widetilde{z}}_{t-1}\right \|^2.
\end{equation}
\vspace{-6pt}
After training the $\psi_t$, we use the trained $\psi_t$ and obtain a better $\mathbf{\widetilde{z}}_{t-1}$ by following:
\begin{equation}\label{eq:pred_next}
 \mathbf{\widetilde{z}}_{t-1} = \sqrt{\frac{\alpha_{t-1}}{\alpha_t}}\widetilde{\mathbf{z}}_t+\left(\sqrt{\frac{1}{\alpha_{t-1}}-1}-\sqrt{\frac{1}{\alpha_t}-1}\right) \cdot \epsilon_\theta(\widetilde{\mathbf{z}}_t,t, \mathbf{c}_{s}, \psi_t).
\end{equation}
With the learning procedure above, the well-optimized $\psi_{t}$ actually preserves the content of the source real image $I_s$ well at timestep $t$ of reverse process, which is important for the next editing stage. The algorithm of the inversion and tuning stages is provided in Algorithm~\ref{alg:alg_twostage}.

\begin{figure}[t]
    \centering
    \includegraphics[width=0.99\linewidth]{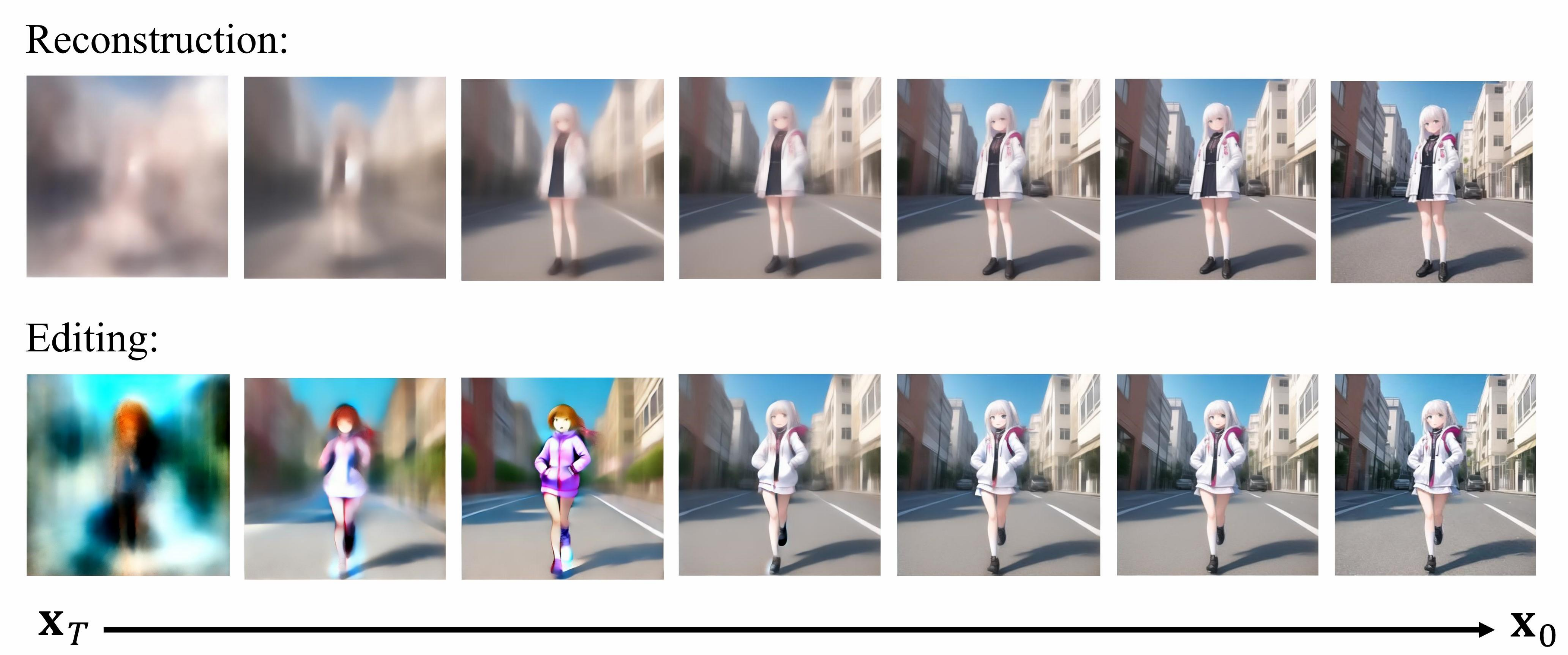}
    \vspace{-5pt}
    \caption{The intermediate predicted $\x_0$ during the sampling process. KV Inversion can ensure the performance of both reconstruction and editing.}
    \label{fig:intermediate_vis}
    \vspace{-14pt}
\end{figure}

\textbf{3) Edit Stage: Using Learned KV Embeddings for Better Content Consistency.}
The overall architecture of the proposed pipeline to perform editing is shown in Fig.~\ref{fig:main_arch}. During each denoising step $t$ of generating the target editing image $I_t$, we also upgrade the self-attention layer in the U-Net to our CP-attn layer with the learned parameters $\psi_{t}$. Specifically, we keep the Query features $Q_{l,t}$ unchanged and use the learned $(\bar{K}_{l,t},\bar{V}_{l,t})$ from the corresponding place of the tuning stage as the new Key and Value features here. Note that $(\bar{K}_{l,t},\bar{V}_{l,t})$ preserve the content of the source image $I_s$. Thus, we can perform attention according to Eq.~\ref{eq:attention} to transfer the contents from $I_s$ to the editing features as shown in Fig.~\ref{fig:intermediate_vis}. We provide the detailed implementation of the method inside the supplementary material.

\vspace{-4pt}

\begin{figure*}[t]
    \centering
    \includegraphics[width=0.99\linewidth]{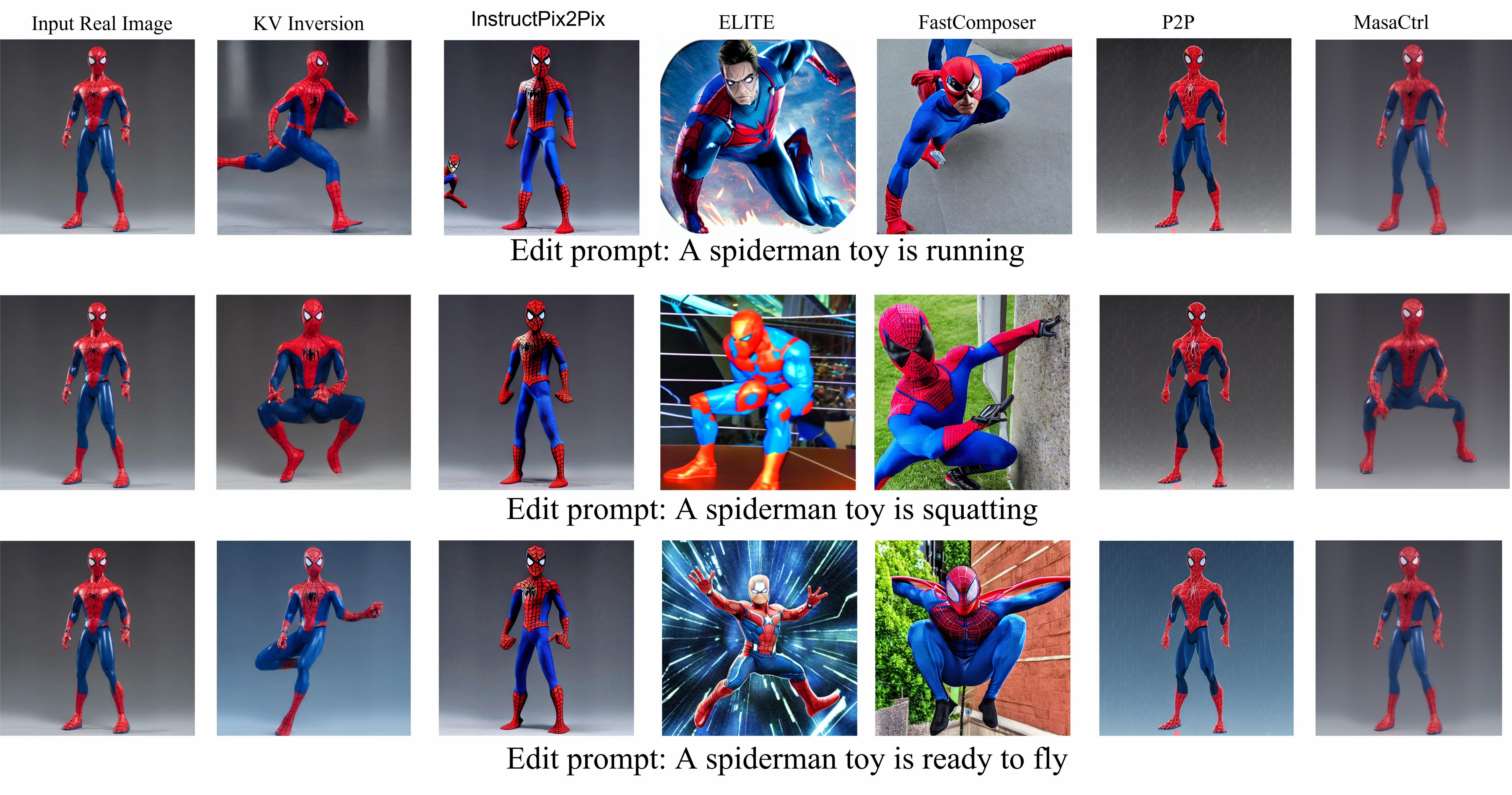}
    \vspace{-12pt}
    \caption{Real image editing results of different editing methods on real human shape images. It is obvious that our KV Inversion can achieve very good action editing performance on humanoid objects, and the semantics of the actions corresponding to its limbs are almost imperceptible to editing.}
    \label{fig:results_human}
    \vspace{-23pt}
\end{figure*}

\begin{figure*}[t]
    \centering
    \includegraphics[width=\linewidth]{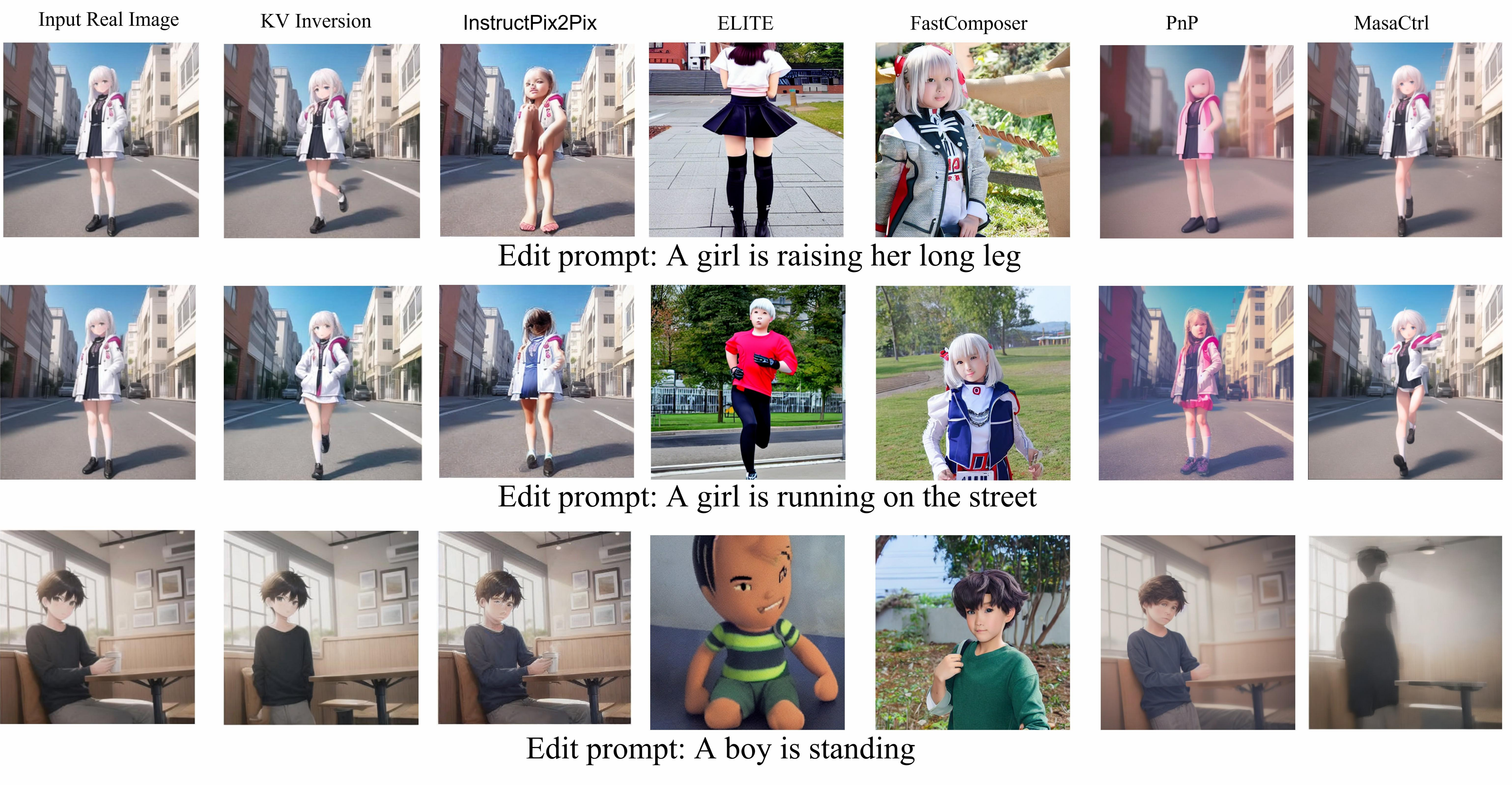}
    \vspace{-23pt}
    \caption{Real image editing results on anime character images. It is clear that our KV inversion method can achieve very good action editing performance on anime characters and their limbs correspond to natural action semantics.}
    \label{fig:real_anime}
    \vspace{-8pt}
\end{figure*}

\vspace{-3pt}
\section{Experiments}
\vspace{-6pt}
We apply the proposed method to the state-of-the-art text-to-image Stable Diffusion~\cite{rombach2022high} model and the anime-style model Anything-V3 with publicly available checkpoints. We focus our experiments on real image action editing. In the inversion, tuning and editing stages, we utilize DDIM sampling~\cite{song2020denoising} with 50 denoising steps, and the classifier-free guidance is set to 7.5. Other hyperparameters may be changed for specific models.
\vspace{-9pt}
\subsection{Comparisons with Other Concurrent Works}
\textbf{On Pretrained Stable Model.}
We mainly compare the proposed KV Inversion to the concurrent text-conditioned image editing methods, including Custorm Diffusion~\cite{kumari2022customdiffusion}, InstructPix2Pix~\cite{brooks2022instructpix2pix}, MasaCtrl~\cite{cao2023masactrl}, P2P~\cite{hertz2022prompt}, PnP~\cite{tumanyan2022plug}, ELITE~\cite{wei2023elite}, FastComposer~\cite{xiao2023fastcomposer}. We use their codes and checkpoints to produce the editing results. 

The editng results are shown in Fig.~\ref{fig:results_real} and Fig.~\ref{fig:results_human}. Our KV Inversion achieves better results in real image editing. As can be seen, action editing for real images is still a very challenging task. Most methods are unable to solve both 1) the resulting edits are not sufficiently similar or even completely different from the source image, and 2) the resulting edits do not match the semantics of the edit prompt in terms of actions.
We see that the proposed KV Inversion solves both of these problems relatively well, and these results demonstrate the effectiveness of the proposed method. Also, unlike methods such as Custorm Diffusion~\cite{kumari2022customdiffusion}, which require fine-tuning of the stable diffusion network itself, our method does not require fine-tuning of the network itself, but only learning some parameters corresponding to the image. In addition, unlike methods such as FastComposer~\cite{xiao2023fastcomposer} and ELITE~\cite{wei2023elite}, which require training on large datasets, we do not need to collect and use large datasets, nor do we need to spend graphics resources and time training on them. The reasons for the failure of the existing methods can be explained by the fact that P2P~\cite{hertz2022prompt} uses the attention map of the source image in the generation of the edited image, thus replicating the original spatial integrity. MasaCtrl~\cite{cao2023masactrl} generates the corresponding action edits, but the reconstruction performance on the real image is still not good enough.

\textbf{On Preatrained Anything-V3 Model.}
We continue to test the effectiveness of our method on the editing of images in the animation domain, specifically using Anything-V3. Fig.~\ref{fig:real_anime} shows the editing results of our method and a number of other methods, including P2P, ELITE, MasaCtrl, InstructPix2Pix, PnP, FastComposer. Note in particular that our method is focused on real image input, and these results in the illustration are edited using the input image provided by the user, rather than given a prompt and then edited during generation, which are two completely different tasks. The proposed KV Inversion is a further demonstration of the generality of the proposed method by retaining the textures and identities of the animated objects in the source image while obtaining the corresponding editing actions.

\begin{figure*}[!t]
    \centering
    \includegraphics[width=0.99\linewidth]{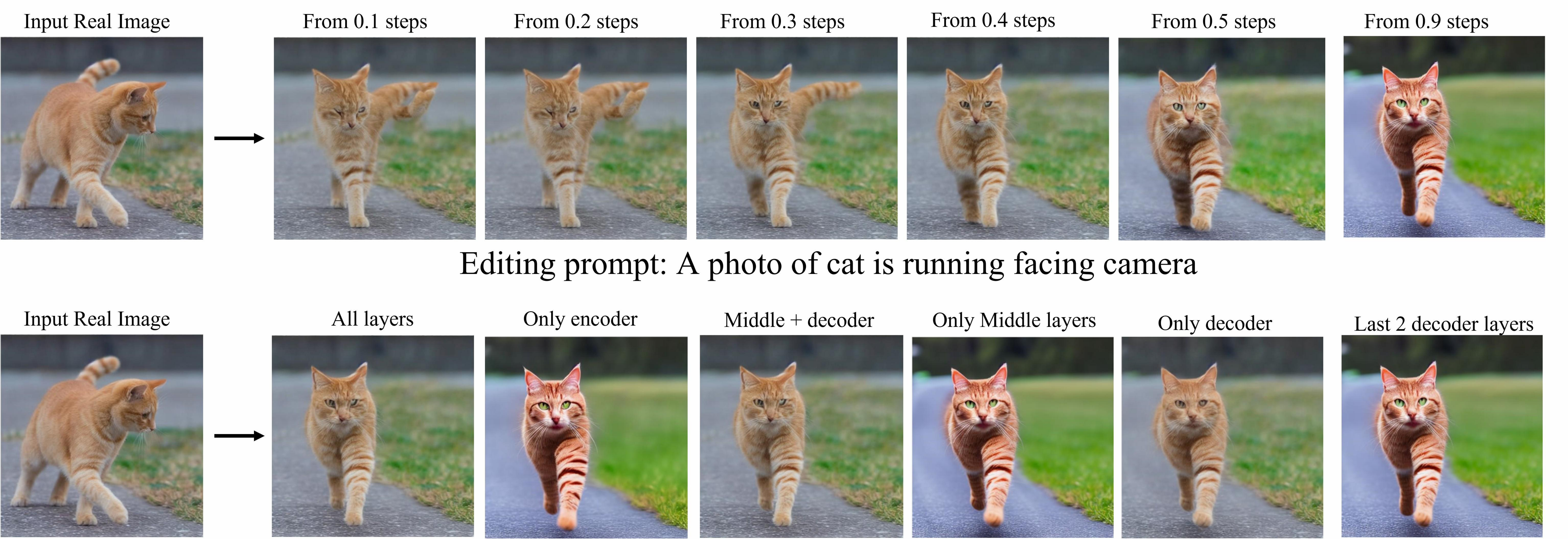}
    \vspace{-4pt}
    \caption{Ablation study of upgrading self-attention layer to CP-attn in different reverse timesteps~(top) and U-Net places~(bottom).}
    \label{fig:step_layer_analysis}
     \vspace{-20pt}
\end{figure*}
\vspace{-8pt}
\subsection{Ablation Study}\label{sec:ablation_study}
\vspace{-3pt}
The effectiveness of the proposed KV Inversion method and the upgrading of self-attention with CP-attn can be demonstrated on top of the comparison results of action editing of real images. 

In order to analyse why the method works, we control the specific parameters for upgrading self-attention to CP-attn, such as the step and position in the U-Net, and observe the effect of the reconstruction and the effect of the editing. From Fig.~\ref{fig:step_layer_analysis} (top), we see that using CP-attn in too few steps (From 0.9 steps) leads to poor performance of the reconstructed image, resulting in an edited result that does not look like the original. Conversely, using CP-attn in too many steps (From 0.1 steps) leads to the edited result too similar to the source image with strange pose, ignoring the requirements of action editing. Similarly, in Fig.~\ref{fig:step_layer_analysis} (bottom) we observe a similar phenomenon when upgrading to CP-attn is performed in different layers of U-Net (Encoder, Middle layer, Decoder). Upgrading to CP-attn in all layers makes the edit texture too similar to the source image. Content preserving performance is poor when only encoder or only middle layers is upgraded to CP-attn, while good reconstruction and editing performance is obtained when only decoder is upgraded to CP-attn.
\vspace{-10pt}
\section{Limitations and Conclusion}
\vspace{-10pt}
Our method also has some obvious problems, the reconstruction performance of our method is good enough, so the problem is not in the tuning stage, but in the editing stage, that is, if the action semantics from the editing prompt that produces an action that is seriously incompatible with the original real image, the editing result will fail. This problem can be solved by having some user provided additional information (such as skeleton joint maps, depth maps, sketch images, segmentation masks, etc.) to control the action~\cite{mou2023t2i, zhang2023adding}, but this is not in line with our original intention of action editing, because providing such additional information is very tedious for the user.

In general, we propose KV Inversion, a method that can achieve highly reconstructed effects and action editing, which can solve two major problems: 1) the edited result can match the corresponding action, and 2) the edited object can keep the texture and identity of the original real image. In addition, our method does not require training U-Net itself, nor does it require scanning large datasets for time-consuming training.
\vspace{-10pt}

\bibliographystyle{splncs04}
\bibliography{egbib}

\end{document}